\documentclass[12pt]{spieman}  
\usepackage{amsmath,amsfonts,amssymb}
\usepackage{graphicx}
\usepackage{setspace}
\usepackage[subfigure]{tocloft}

\usepackage[vlined, ruled, boxed]{algorithm2e}
\SetKwInput{kwPreprocessing}{Preprocessing}
\usepackage{subfigure}

\title{Clustered Multitask Nonnegative Matrix Factorization for Spectral Unmixing of Hyperspectral Data}

\author[a]{Sara Khoshsokhan}
\author[a,*]{Roozbeh Rajabi}
\author[a,b]{Hadi Zayyani}
\affil[a]{Qom University of Technology, Faculty of Electrical and Computer Engineering, Khodakaram Blvd, Qom, Iran, 1519-37195}
\affil[b]{Shiraz University, School of Electrical and Computer Engineering, Zand Street, Shiraz, Iran, 71348-51154}

\cftpagenumbersoff{figure}
\cftpagenumbersoff{table} 
\begin{document}
\maketitle

\begin{abstract}
In this paper, the new algorithm based on clustered multitask network is proposed to solve spectral unmixing problem in hyperspectral imagery. In the proposed algorithm, the clustered network is employed. Each pixel in the hyperspectral image considered as a node in this network. The nodes in the network are clustered using the fuzzy c-means clustering method. Diffusion least mean square strategy has been used to optimize the proposed cost function. To evaluate the proposed method, experiments are conducted on synthetic and real datasets. Simulation results based on spectral angle distance, abundance angle distance and reconstruction error metrics illustrate the advantage of the proposed algorithm compared with other methods.
\end{abstract}


{\noindent \footnotesize\textbf{*}Roozbeh Rajabi,  \linkable{rajabi@qut.ac.ir} }

\begin{spacing}{2}   

\section{Introduction}
\label{sect:intro}  
One of the noteworthy remote sensing techniques is hyperspectral imaging. Hyperspectral images provide rich spectral information, because the sensors contain hundreds of spectral channels with higher spectral resolution than multispectral cameras. For example, the images, generating from the Airborne Visible Infra-Red Imaging Spectrometer (AVIRIS), have 224 bands. One of the problems in hyperspectral data is presence of mixed pixels \cite{IJRS12_ReviewSpectralUnmixing}. In the scene, pixels containing a single material are called pure pixels and otherwise they are called mixed pixels \cite{Agg16}. Each pixel is composed of a set of materials called endmembers. The corresponding fraction of an endmember in that pixel is named fractional abundance \cite{Miao07}. Spectral unmixing (SU)  methods decompose a reflectance spectrum into a set of endmember spectra \cite{Mei11} and their abundance fractions. The most common mixing model for hyperspectral data is linear mixing model (LMM) in which it is supposed that the recorded reflectance of a particular pixel is a linear combination of its endmembers. In contrast with nonlinear (intimate) models \cite{JARS17_Nonlinear}, LMM features simplicity, acceptable efficiency and low computational complexity. Because of these features, there exists many works that exploit the LMM to solve unmixing problem. Examples of such works are structured sparse method \cite{ISPRS14_Structured}, minimum volume simplex analysis (MVSA) \cite{TGRS15_MVSA} that is a classic unmixing algorithm based on a minimum volume simplex, improved discrete swarm intelligence \cite{JARS16_Swarm}, conantroppy maximization using alternating direction method of multipliers (ADMM) \cite{TGRS17_Correntropy} and stacked nonnegative sparse autoencoders \cite{GRSL18_Stacked} that is a special case of artificial neural network (ANN) and has the ability to extract deep robust features.

Nonnegative matrix factorization (NMF) \cite{Paatero94,Lee99} is one of the practical methods of spectral unmixing, which decomposes the data into two nonnegative matrices. Recently, this basic method was developed by adding constraints, such as the minimum volume constrained NMF (MVC-NMF) method \cite{Miao07}, graph regularized NMF (GNMF) \cite{Rajabi13}, NMF with local smoothness constraint (NMF-LSC) \cite{Yang15}, multilayer NMF (MLNMF) \cite{Rajabi15}, region based structured NMF \cite{JSTARS17_RegionBasedStructuredNMF} and NMF based framework for hyperspectral unmixing using prior knowledge (NMFupk) \cite{Optical12_NMFupk}. Sparsity is one of the constraints for improving performance of NMF algorithm that is applied to the NMF cost function using $ L_ {q} $ regularizers \cite{Qian11}, or using double reweighted sparse regression and total variation \cite{wang17}. Since the number of endmembers present in each mixed pixel is small in comparison with the number of total endmembers, the problem becomes sparse \cite{Irodache10}. $ L_ {1/2} $-NMF unmixing algorithm is developed by applying $ L_ {q} $ regularization term into NMF cost function to enforces the sparsity of endmember abundances \cite{Qian11}.

Recently, distributed strategy gained a lot of interest in many areas \cite{Chen14}. Diffusion least mean square (LMS) solution has been used to solve distributed problem \cite{Wen13}. In a distributed optimization problem, there is a network with three types of structures: 1) a single-task network, that nodes estimate a common unknown and optimum vector, 2) a multitask network, which each node estimate its own optimum vector and 3) a clustered multitask network, which contains clusters that in each of clusters there is a common optimum vector that should be estimated \cite{Chen14}. In the hyperspectral image, a network of pixels can be considered to model the SU problem as a distributed one \cite{khosh17,JSTARS19_DistributedUnmixing}. Here, we have used clustered multitask network to solve SU problem \cite{ICSPIS18_ClusteredMultitask}. Using clustered multitask network, only the neighborhood information of spectrally similar mixed pixels (those are in the same cluster) will be used in the proposed cost function. In order to generate a clustered network, we used the FCM clustering method on spectral features of hyperspectral data. Then unmixing problem has been solved as a clustered multitask network using information of nodes in neighborhood and clusters in addition to sparsity constraint.

This paper is organized as follows. In section 2, we introduce the clustered multitask network and proposed unmixing algorithm. Section 3 includes evaluation criteria, experiments on synthetic and real datasets and comparison with other methods. Section 4 gives conclusions and future work.

\section{Spectral Unmixing Using Clustered Multitask Network}

\begin{figure}
	\centering
	\includegraphics[width=6in]{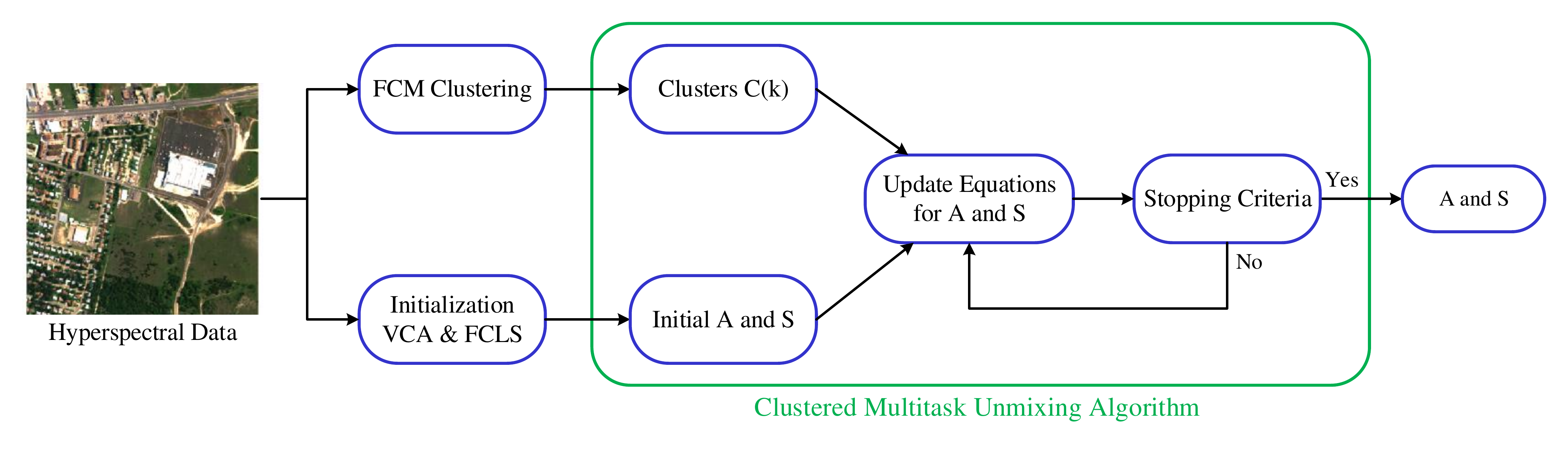}
	\caption{Block diagram of the proposed method.}
	\label{blockDiagram}
\end{figure}

In this section, a new method that utilizes clustering of pixels and neighborhood information is proposed. First, we will express linear mixing model in subsection 2.1, then the distributed algorithm, cost functions and optimization procedure are formulated in 2.2. Finally, the overall algorithm to solve SU problem is presented in 2.3. The proposed method to solve spectral unmixing problem has been summarized in \figurename{~\ref{blockDiagram}}.

\subsection{Linear Mixing Model}
\label{sect:title}

To solve the SU problem, we focus on a simple and applicable model named Linear Mixing Model (LMM). In this model, observations are a linear combination of endmembers and their fractional abundances in each pixel. Mathematically, this model for pixel $k$ in the hyperspectral image is described as:
\begin{equation}
\mathbf{y}_k=\mathbf{A}\mathbf{s}_k+\mathbf{v}_k
\end{equation}
where $\mathbf{y}_k$ is an $L\times1$ observed data vector, $\mathbf{A}$ is the $L\times c$ signature matrix, $\mathbf{s}_k=[s_k(1),s_k(2),..,s_k(c)]^T$ is the $c\times1$ fractional abundance vector and $\mathbf{v}_k$ is assumed as a $L\times1$ additive noise vector of $k$-th pixel of the image, when $c$, $L$ and $N$ denote the number of endmembers, bands and pixels, respectively.

In the SU problem, fractional abundance vectors have two constraints in each pixel, abundance sum to one constraint (ASC) and abundance nonnegativity constraint (ANC) \cite{Ma14}, which are as follows, for $c$ endmembers in a scene. 
\begin{equation}
\sum\limits_{n=1}^c s_{k}(n)=1
\end{equation}
\begin{equation}
s_{k}(n)\geq0,n=1,...,c
\end{equation}
where $s_{k} (n)$ is the fractional abundance of the $n$-th endmember in the $k$-th pixel of the image.

\subsection{Distributed Cost Functions and Optimization}
As explained earlier, we can consider three types of networks containing single task, multitask and clustered multitask networks. First, $N$ nodes are considered in a clustered multitask network and a optimum vector at node $k$ is estimated. A global cost function using LMS, $J^{global}(\mathbf{s}_1,\mathbf{s}_2,...,\mathbf{s}_N)$, defined as follows:
\begin{equation}
\label{eq: 4}
J^{global} (\mathbf{s}_1,\mathbf{s}_2,...,\mathbf{s}_N)=\sum\limits_{k=1}^N \mathrm{E}\{|\mathbf{y}_k -\mathbf{A} \mathbf{s}_k|^{2}\}
\end{equation}
where $\mathrm{E}$ is the expectation operator. Then, to minimize the cost function, the following equation is obtained, using the iterative steepest-descent solution \cite{Cattivelii10}:
\begin{equation}
\label{eq: 6}
\mathbf{s}_k^{i}=\mathbf{s}_k^{i-1}+\mu \sum\limits_{k=1}^N \mathbf{A}^T (\mathbf{y}_k-\mathbf{A}\mathbf{s}_k) 
\end{equation}
where $\mu>0$ is a step-size parameter, and the algorithm make small jumps, using an optimum value of $\mu$. This optimum value causes stability and depends on the cost function. The algorithm will diverge with a too large value of $\mu$, and will take a long time to converge with a too small value. $i$ is iteration number.

In equation (\ref{eq: 6}) the neighborhood information has not been used yet. In a distributed network, information from neighboring nodes are used to improve accuracy. In this article, we utilize the squared Euclidean distance \cite{Chen14}:
\begin{equation}
\label{eq: 8}
\Delta(\mathbf{s}_k,\mathbf{s}_l)=||\mathbf{s}_k-\mathbf{s}_l||^2
\end{equation}
And then, the $L_{q}$ regularizer for sparsity constraint is used \cite{Qian11}:
\begin{equation}
\label{eq: 9}
||\mathbf{s}_k||_{q}=\big( \sum\limits_{n=1}^c \mathbf{s}_{k}^{q} (n)\big)^{1/q}
\end{equation}
Note that, the solution determined from global cost function, need to have access to information over cluster of the node, but the nodes can be considered to have availability only to information of its neighbors and the nodes of in the same cluster. Thus, for solving this problem, the following local cost function is defined, using LMS and adding the (\ref{eq: 8}) and (\ref{eq: 9}) constraints:
\begin{equation}
\label{eq: 10}
J^{local} (\mathbf{s}_k)= \mathrm{E}\{|\mathbf{y}_k-\mathbf{A} \mathbf{s}_k|^{2}\}
+\eta \sum\limits_{l\in \mathcal{N}_k \cap C(k)} \rho_{kl}  ||\mathbf{s}_{k}-\mathbf{s}_{l}||^2
+ \lambda ||\mathbf{s}_k||_{q}
\end{equation}
where the $\mathcal{N}_k$ shows nodes that are in the neighborhood of node $k$, the node that exists in the cluster $C(k)$. $\eta>0$  denotes a regularization parameter \cite{Chen14}, that controls the effect of neighborhood term, $\lambda$ is a scalar value that weights the sparsity function \cite{Qian11}, and the nonnegative coefficients $\rho_{kl}$ are normalized spectral similarity which are obtained from correlation of data vectors \cite{Chen14}:
\begin{equation}
\label{eq: lambda}
\lambda = \frac{1}{\sqrt{L}} \sum\limits_{l}  \frac{\sqrt{N}-||\mathbf{y}_l||_1/||\mathbf{y}_l||_2}{\sqrt{N-1}}
\end{equation}
where $\mathbf{y}_l$ denotes the $l$th band of the hyperspectral image, in this equation.
\begin{equation}
\label{eq: rho}
\rho_{kj} = \frac{\theta (\mathbf{y}_k,\mathbf{y}_j)}{\sum\limits_{l \in\mathcal{N}_k^-} \theta (\mathbf{y}_k,\mathbf{y}_l)}
\end{equation}
where $\mathcal{N}_k^-$ include neighbors of node $k$ except itself, and $\theta$ is computed as \cite{Chen14}:
\begin{equation}
\label{eq: theta}
\theta (\mathbf{y}_k,\mathbf{y}_j) = \frac{\mathbf{y}_k^T \mathbf{y}_j}{||\mathbf{y}_k|| ||\mathbf{y}_j||}
\end{equation}

Now, minimizing the cost function of (\ref{eq: 10}), using steepest-descent algorithm, results to:
\begin{equation}
\label{eq: s}
\mathbf{s}_k^{i+1} = \mathbf{s}_k^{i}+ \mu \mathbf{A}^T (\mathbf{y}_k-\mathbf{A}\mathbf{s}_k)
- \mu \eta \sum\limits_{l\in \mathcal{N}_k\cap C(k)} \rho_{kl} (\mathbf{s}_l^{i}-\mathbf{s}_k^{i})
- \mu \lambda \frac{\big( \mathbf{s}_k^{i}\big) |\mathbf{s}_k^{i}|^{q-2}}{||\mathbf{s}_k^{i}||_{q}^{q-1}}
\end{equation}

Hence, this recursive equation can be used to update fractional abundance vectors in the SU problem.
\subsection{Proposed Algorithm}
Similar to the NMF algorithm, the least mean square error should be minimized with respect to the signatures and abundances matrices, subject to the non-negativity constraint \cite{Lee01}. So, the following equation is denoted, using matrix notation:
\begin{equation}
\label{eq: 21}
\min\limits_{\mathbf{S},\mathbf{A}>0} ||\mathbf{Y}-\mathbf{A}\mathbf{S}||_F^2
\end{equation}
where $\mathbf{A}$ and $\mathbf{S}$ are the $L\times c$ signature and $c\times N$ fractional abundances matrices, respectively, and Y denotes the $L\times N$ Hyperspectral data matrix. Then, based on described equations of the sparsity constrained distributed unmixing, the neighborhood and sparsity terms are added to (\ref{eq: 21}) as follows:
\begin{equation}
\label{eq: 22}
||\mathbf{Y}-\mathbf{A}\mathbf{S}||_F^2 + \eta \sum \limits_{k=1}^N \sum\limits_{j\in \mathcal{N}_k \cap C(k)} \rho_{kj} ||\mathbf{s}_j - \mathbf{s}_k||^2+ \lambda \sum \limits_{k=1}^N ||\mathbf{s}_k||_{q}
\end{equation}
This cost function is minimized with respect to $\mathbf{A}$, using multiplicative update rules \cite{Lee01}, then recursive equation of signature matrix is obtained as:
\begin{equation}
\label{eq: A}
\mathbf{A}^{i+1}=\mathbf{A}^i*\frac{\mathbf{Y}\mathbf{S}^T} {\mathbf{A}\mathbf{S}\mathbf{S}^T} 
\end{equation}
And the recursive equation of fractional abundances has been obtained already in accordance with (\ref{eq: s}) as follows:
\begin{equation}
\label{eq: s1}
\mathbf{s}_k^{i+1} = P^+\Big(\mathbf{s}_k^{i}+ \mu \mathbf{A}^T (\mathbf{y}_k-\mathbf{A}\mathbf{s}_k)
- \mu \eta \sum\limits_{l\in \mathcal{N}_k\cap C(k)} \rho_{kl} (\mathbf{s}_l^{i}-\mathbf{s}_k^{i})- \mu \lambda \frac{\big( \mathbf{s}_k^{i}\big) |\mathbf{s}_k^{i}|^{q-2}}{||\mathbf{s}_k^{i}||_{q}^{q-1}} \Big)
\end{equation}
where $P^+$ operator projects vectors onto a simplex, that adopt the ASC and ANC constraints for abundance vectors \cite{Chen11}. To initialize the matrices, random initialization and VCA-FCLS method can be used. Since random values may obtain local optimums, VCA-FCLS initialization method has been used in this paper. Another significant point in implementation of the algorithm is stopping criteria. This approach will be stopped until the maximum number of iteration ($T$), or the following stopping criteria is reached.
\begin{equation}
\label{eq: stop}
||J_{new}-J_{old}||<\epsilon
\end{equation}
where $J_{new}$ and $J_{old}$ are cost function values for two consecutive iterations and $\epsilon$ has been set to $10^{-8}$ in our experiments. Now, the proposed approach is summarized in Algorithm 1.
\begin{algorithm}
	\SetKwInOut{Input}{input}
	\SetKwInOut{Output}{output}
	\SetKwInput{Initialization}{Initialisation}
	\Input{Hyperspectral data matrix ($\mathbf{Y}$)\\
		Parameters: $C$,$N$,$c$,$L$,$q$,$\mu$ and $\eta$,}
	\Output{Estimated fractional abundance and signature matrices ($\mathbf{S}$ and $\mathbf{A}$),}
	\kwPreprocessing{Clustering $\mathbf{Y}$ using FCM algorithm to $C$ clusters, determines $C(k)$, $k=1,...,N$,}
	\Initialization{Initialise the $\mathbf{A}$ and $\mathbf{S}$ matrices by random matrices or the outcome of VCA-FCLS algorithm. Compute  $\rho$ values from (\ref{eq: rho}),}
	\While{the maximum number of iteration ($T$) or stopping criteria in (\ref{eq: stop}) has been reached,}
	{
		a. Update $\mathbf{A}$, using  (\ref{eq: A})\;
		b. Update $\mathbf{s}_k$ for all pixels, by applying  (\ref{eq: s1})\;
		c. Adopt $P^+$ operator for ASC and ANC constraints\;
		end}
	\caption{Hyperspectral Unmixing Based on Clustered Multitask Networks}
	\label{algorithm}
\end{algorithm}

\section{Experiments and Results}

In this section, for quantitative evaluation, two common performance metrics: spectral angle distance (SAD) and abundance angle distance (AAD) \cite{Miao07} are used. They are defined as:

\begin{equation}
\label{eq: 25}
\text{SAD}= \cos^{-1} \Big(\frac{\mathbf{a}^T \hat{\mathbf{a}}}{||\mathbf{a}|| ||\hat{\mathbf{a}}||}\Big)
\end{equation}

\begin{equation}
\label{eq: 26}
\text{AAD}= \cos^{-1} \Big(\frac{\mathbf{s}^T \hat{\mathbf{s}}}{||\mathbf{s}|| ||\hat{\mathbf{s}}||}\Big)
\end{equation}
where $\hat{\mathbf{a}}$ is the estimation of spectral signature vectors and $\hat{\mathbf{s}}$ is the estimation of fractional abundance vectors.

Another criterion that is used for quantitative evaluation is reeconstruction error (RE) that is defined as follows \cite{GRSL18_Stacked}:

\begin{equation}
\label{eq:RE}
\text{RE}= \frac{1}{n}\sum_{i=1}^{n}\sqrt{||\mathbf{y}_i-\mathbf{\hat{y}}_i||_{2}^{2}}
\end{equation}
where $\mathbf{y}_i$ and $\mathbf{\hat{y}}_i$ are the original and reconstructed pixels and $n$ is the total number of pixels in the hyperspectral scene.

\begin{figure}[!t]
	\centering
	\includegraphics[width=2.5in]{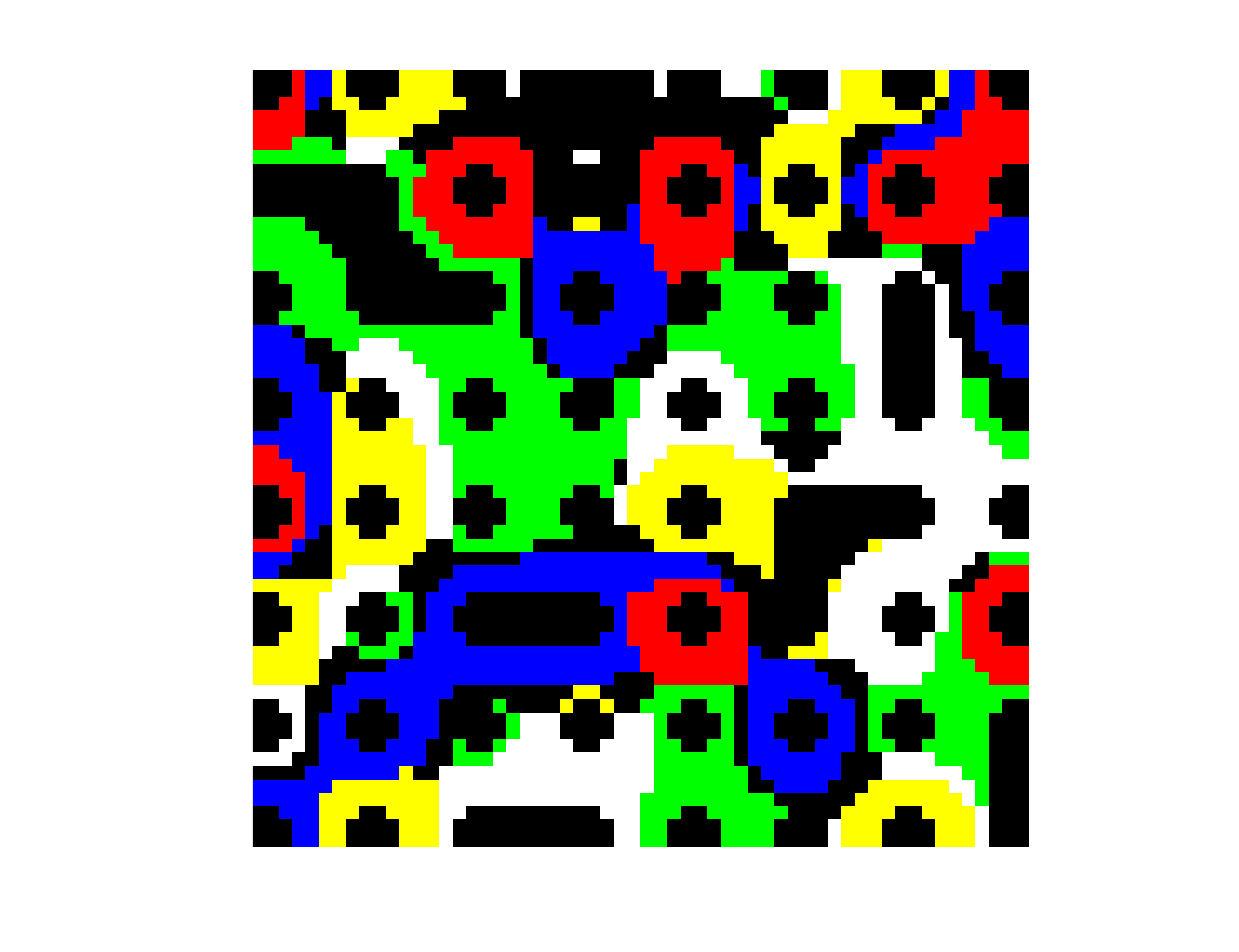}
	\caption{FCM Clustering of synthetic dataset.}
	\label{clus}
\end{figure}
\begin{figure}
	\centering
	\includegraphics[width=2.25in]{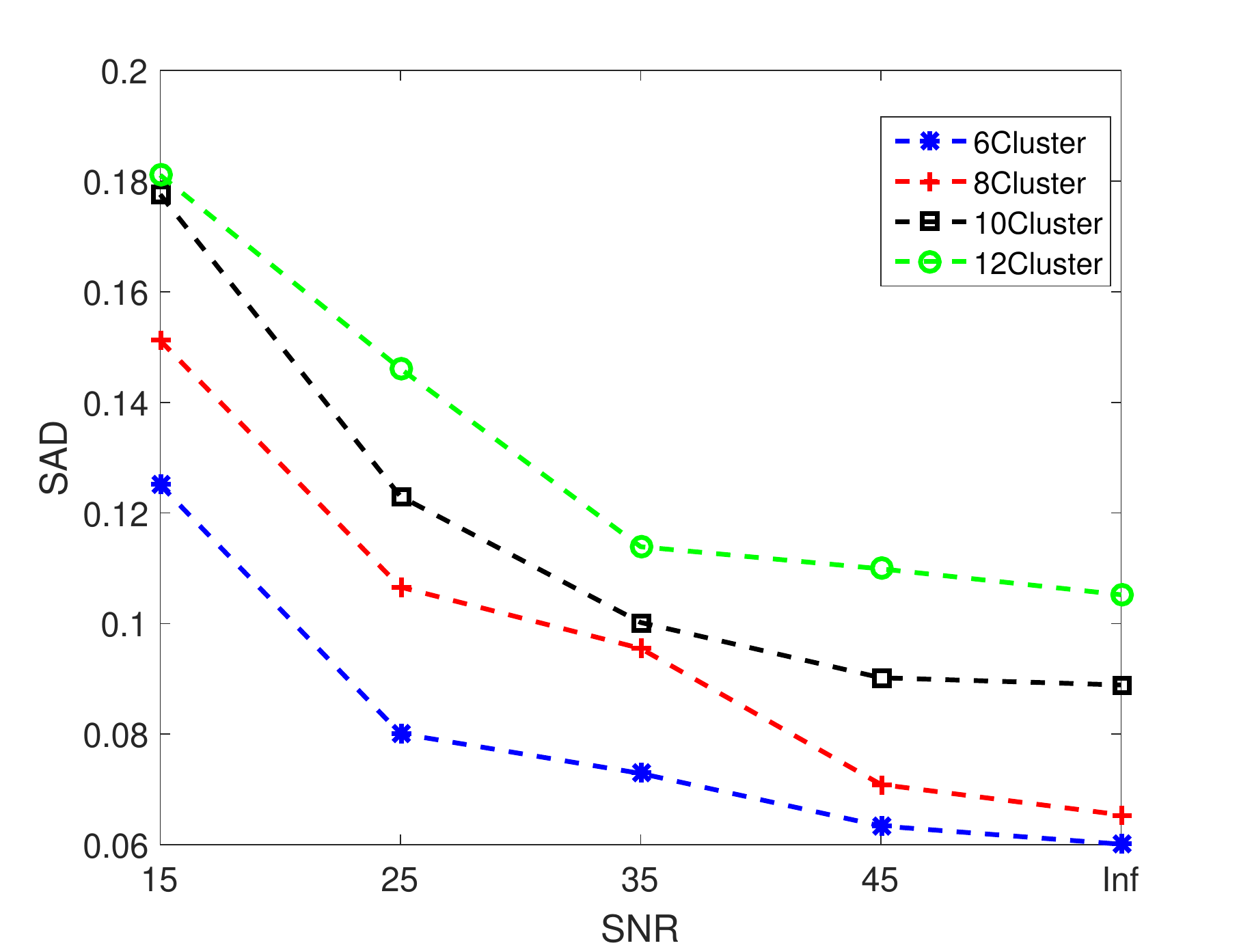}
	\caption{The SAD performance metric of the proposed algorithm applied on synthetic dataset for 6 endmembers, with different number of clusters and using VCA-FCLS initialization.}
	\label{cl}
\end{figure}
\begin{figure}[!t]
	\centering
	\includegraphics[width=6.5in]{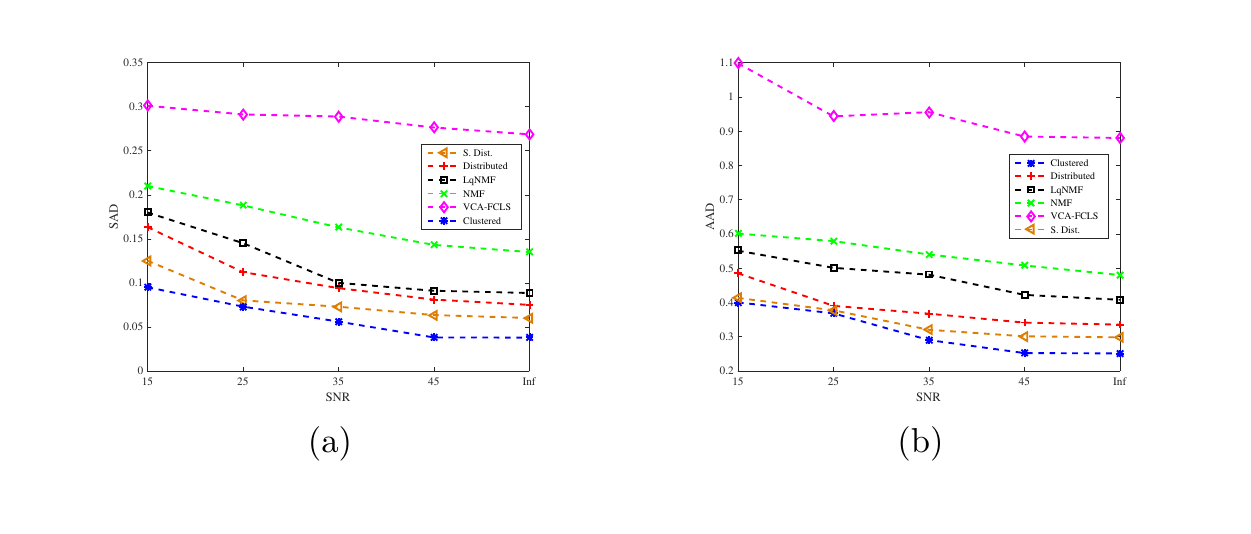}
	\caption{The (a) SAD and (b) AAD performance metric of six different  algorithm applied on synthetic dataset for six endmembers, using VCA-FCLS initialization.}			\label{fig:syntheticSAD_AAD}
\end{figure}
\begin{figure}[!htbp]
	\centering
	\includegraphics[width=5in]{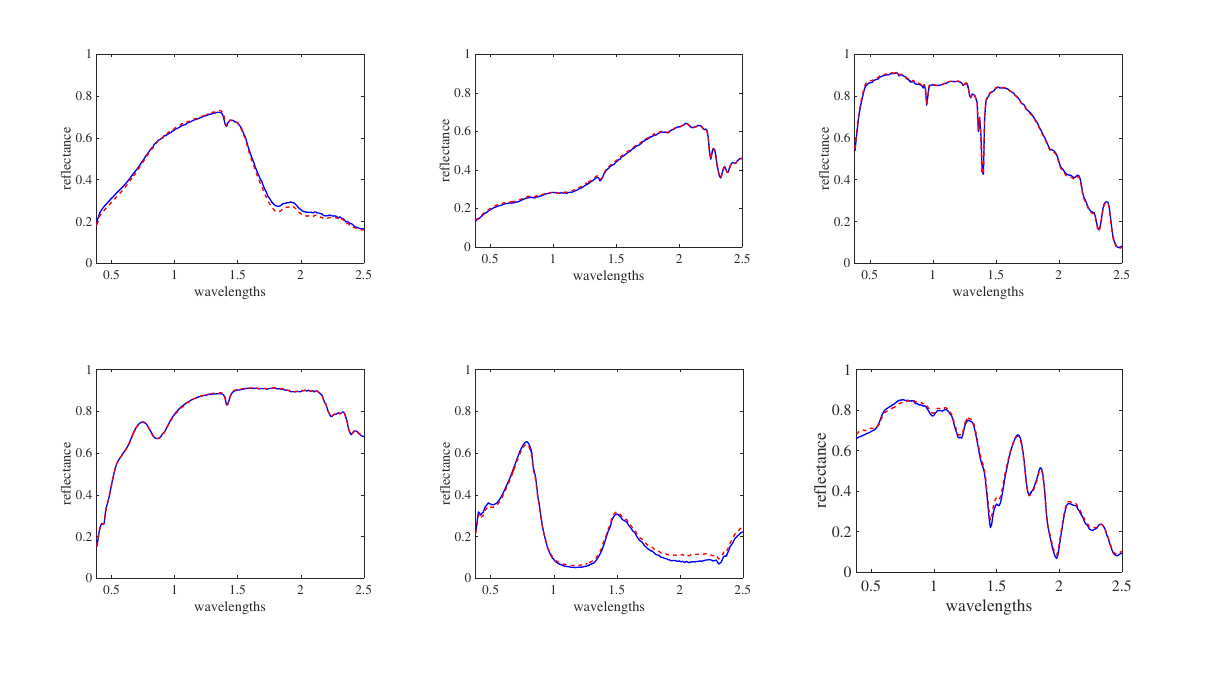}
	\caption{Original spectral signatures (blue solid lines) and estimated signatures of proposed algorithm (red dashed lines) versus wavelengths ($\mu m$), on synthetic data and using VCA-FCLS initialization with SNR=25dB.}
	\label{syn}
\end{figure}

\begin{figure}
	\centering
	\includegraphics[width=5in]{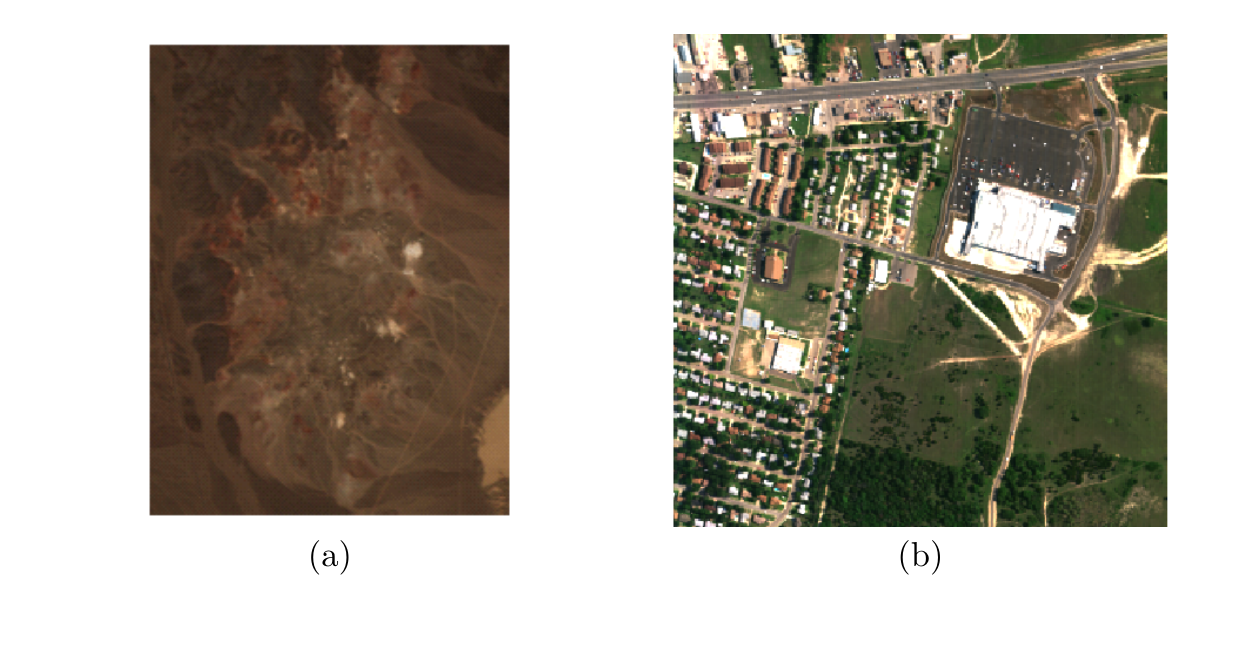}
	\caption{Pseudo color image of (a) AVIRIS Cuprite data scene. The bands used as RGB channels are bands (40,20,10) of original 224 bands image and (b) HYDICE Urban data scene. The bands used as RGB channels are bands (49,35,18) of original 210 bands image.}
	\label{fig:colorReal}
\end{figure} 

In this article, the proposed algorithm is applied on synthetic and real datasets. First, the proposed algorithm has been applied on synthetic data. to generate this dataset, six signatures of USGS library \cite{Clark07} have been selected randomly, using a 7$\times$7 low pass filter and containing no pure pixels \cite{Miao07}. Then, the zero mean Gaussian noise with 5 different levels of SNR has been added to generated data, and performance metrics have been computed by averaging 20 Monte-Carlo runs. In \figurename{~\ref{clus}}, the FCM clustering of this dataset is illustrated. Then, to choose the best number of clusters in our experiments, the SAD performance metric has been evaluated, and then according to \figurename{~\ref{cl}}, the best number of clusters has been set to 6, that is equal to number of endmembers. Also, values of $\mu$ and $\eta$ has been considered equal to 0.02 and 0.1, respectively \cite{Chen14}, and $q=2$, to gain the best results.

Then the proposed algorithm and some other algorithms: VCA-FCLS, NMF, $L_{1/2}$-NMF, distributed unmixing and sparsity constrained distributed unmixing, has been applied on the generated synthetic dataset. The comparison of performance metrics of this six different methods has been shown in \figurename{~\ref{fig:syntheticSAD_AAD}} (a) and (b), where the metrics of proposed algorithm is star-dashed line and excels other methods. \figurename{~\ref{syn}} shows the original and estimated spectral signatures for 6 endmembers, when the SNR is set to $25dB$.

\begin{figure}
	\centering
	\includegraphics[width=3in]{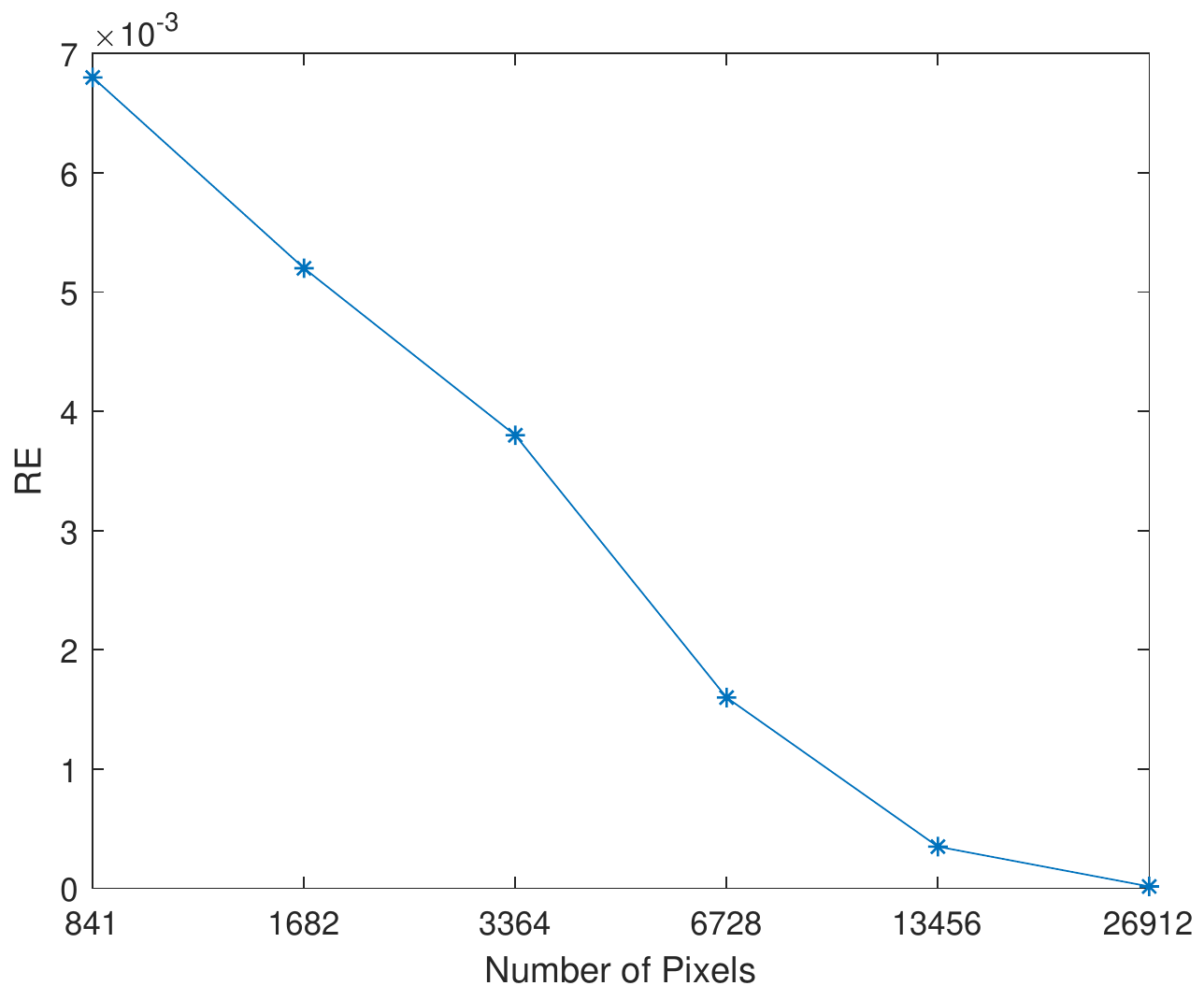}
	\caption{The RE performance metric of the proposed algorithm applied on synthetic dataset for 6 endmembers, with different number of pixels and using VCA-FCLS initialization.}
	\label{fig:synNumberPixels}
\end{figure}

Another experiment has been conducted to show the performance of the proposed algorithm for different number of pixels in the synthetic hyperspectral image. Other parameters have been set like the previous experiments. The RE metric defined in \eqref{eq:RE} is used in this experiment. \figurename{~\ref{fig:synNumberPixels}} illustrates the results of this experiment. As it can be seen in this figure, the proposed algorithm performs better when the size of hyperspectral data grows. This emphasizes that with increase of the training data size, the algorithm converges to a solution more robustly.

In the next two experiments, the proposed algorithm has been applied on real datasets: AVIRIS Cuprite and HYDICE Urban. AVIRIS Cuprite dataset is a hyperspectral data captured by the AVIRIS sensor over Cuprite, Nevada. This sensor covers wavelengths from 0.4$\mu$$m$ to 2.5$\mu$$m$ in 224 channels \cite{Green98}. 188 bands of these 224 bands are used in the experiments and the other bands (covering bands 1, 2, 104-113, 148-167, and 221-224) have been removed which are related to water-vapor absorption or low SNR bands. \figurename{~\ref{fig:colorReal}} (a) illustrates a pseudo color image of this dataset.

In HYDICE Urban dataset, There are 210 bands, that covers wavelengths from 0.4$\mu$$m$ to 2.5$\mu$$m$. After removing water-vapor absorption or low SNR bands (including 1-4, 76, 87, 101-111, 136-153, and 198-210), 162 bands are used in the experiments. There are 4 distinguished materials in HYDICE Urban image: asphalt, roof, tree and grass \cite{He16}. \figurename{~\ref{fig:colorReal}} (b) illustrates a pseudo color image of this real dataset.

\begin{figure}[h]
	\centering
	\includegraphics[width=6.5in]{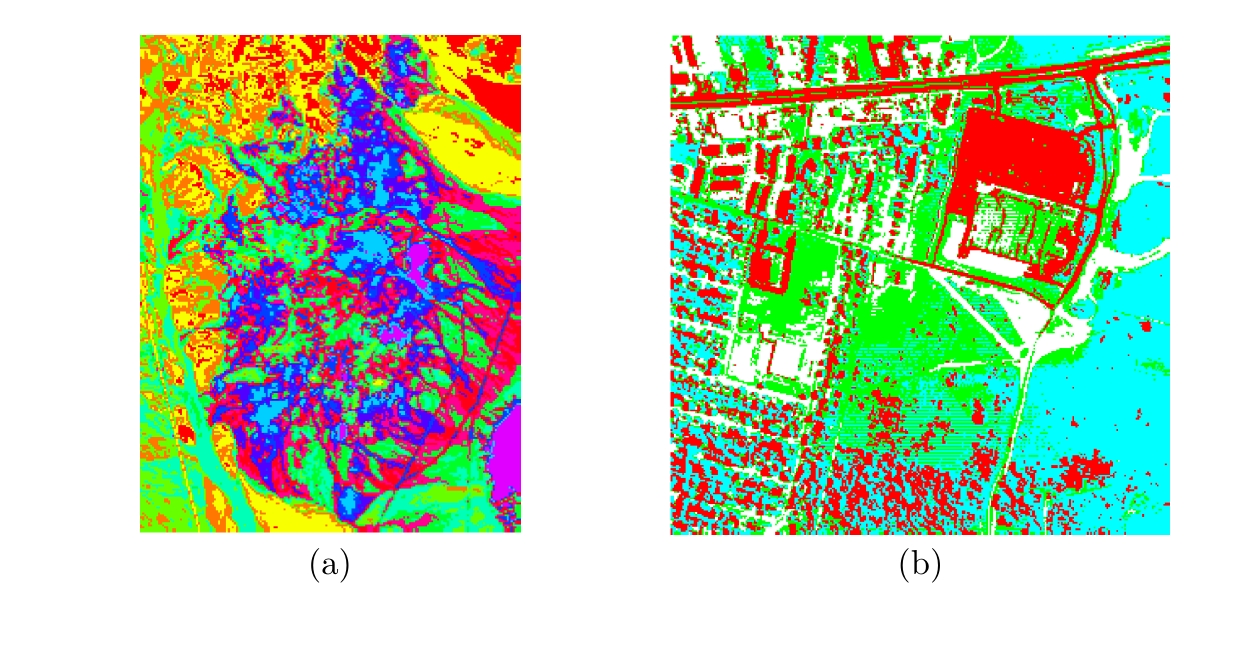}	
	\caption{FCM Clustering of (a) AVIRIS Cuprite and (b) HYDICE Urban datasets.}
	\label{realClustering}
\end{figure}

\begin{figure}[h]
	\centering
	\includegraphics[width=6.5in]{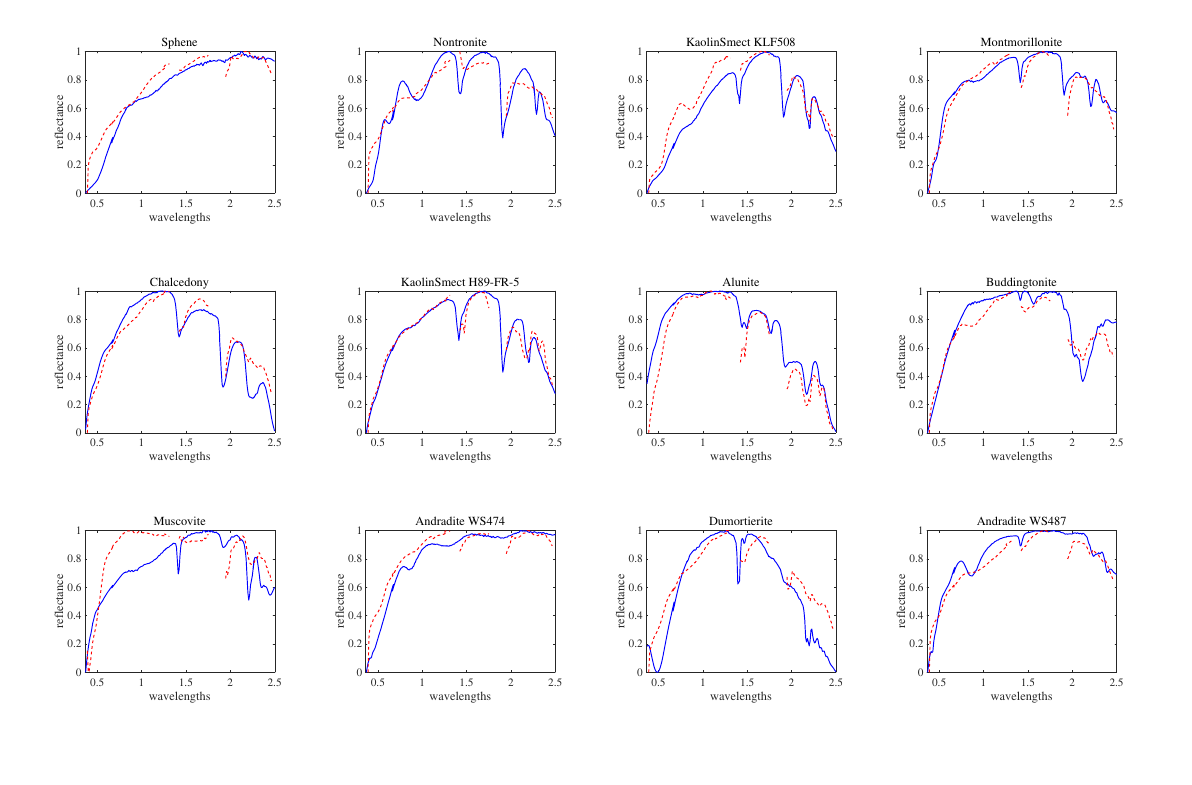}	
	\caption{Original spectral signatures (blue solid lines) and estimated signatures of proposed algorithm (red dashed lines) versus wavelengths ($\mu m$), on AVIRIS Cuprite dataset and using VCA-FCLS initialization.}
	\label{real}
\end{figure}

\begin{figure}[h]
\centering
\includegraphics[width=6.5in]{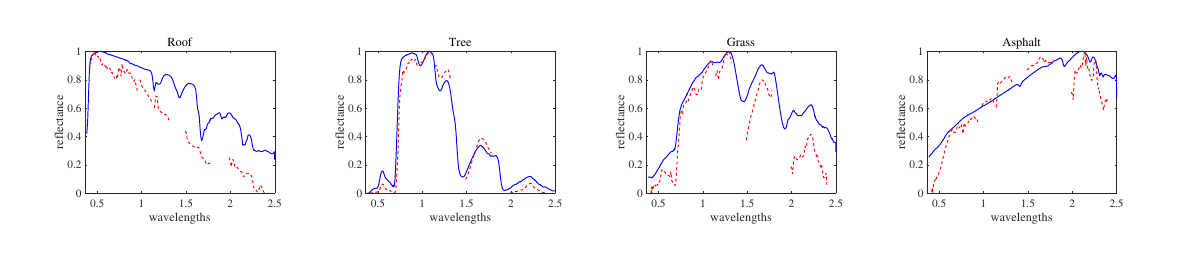}
	
	\caption{Original spectral signatures (blue solid lines) and estimated signatures of sparsity constrained distributed unmixing (red dashed lines) versus wavelengths ($\mu m$), on HYDICE Urban dataset and using VCA-FCLS initialization.}
	\label{realhy}
\end{figure}

\begin{figure}

	\centering
\includegraphics[width=140mm]{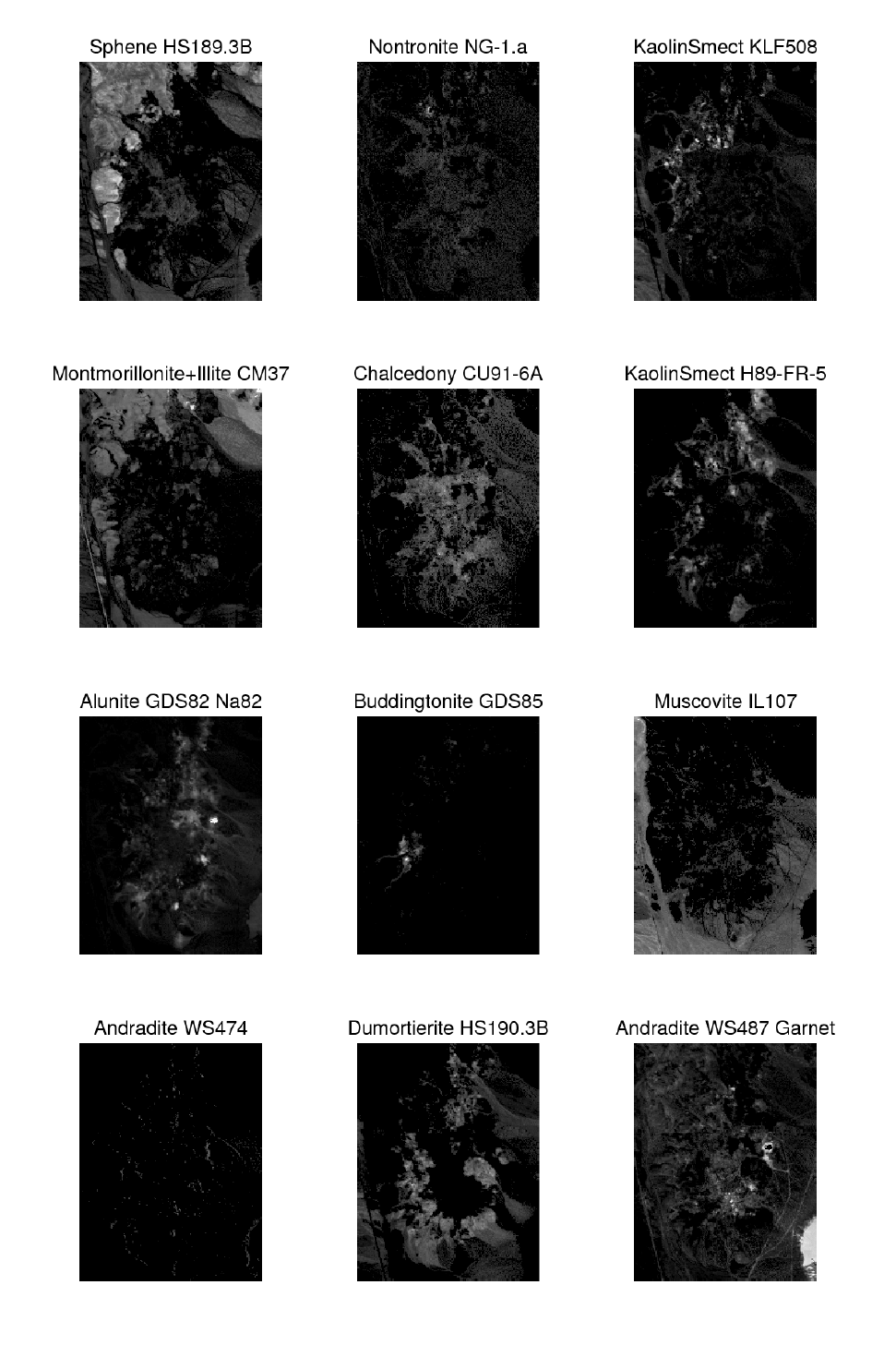}

	\caption{The abundance fraction maps of the proposed algorithm applied on AVIRIS Cuprite dataset.}
	\label{CupriteAbundance}
\end{figure}

\begin{figure}

	\centering
	\includegraphics[width=125mm]{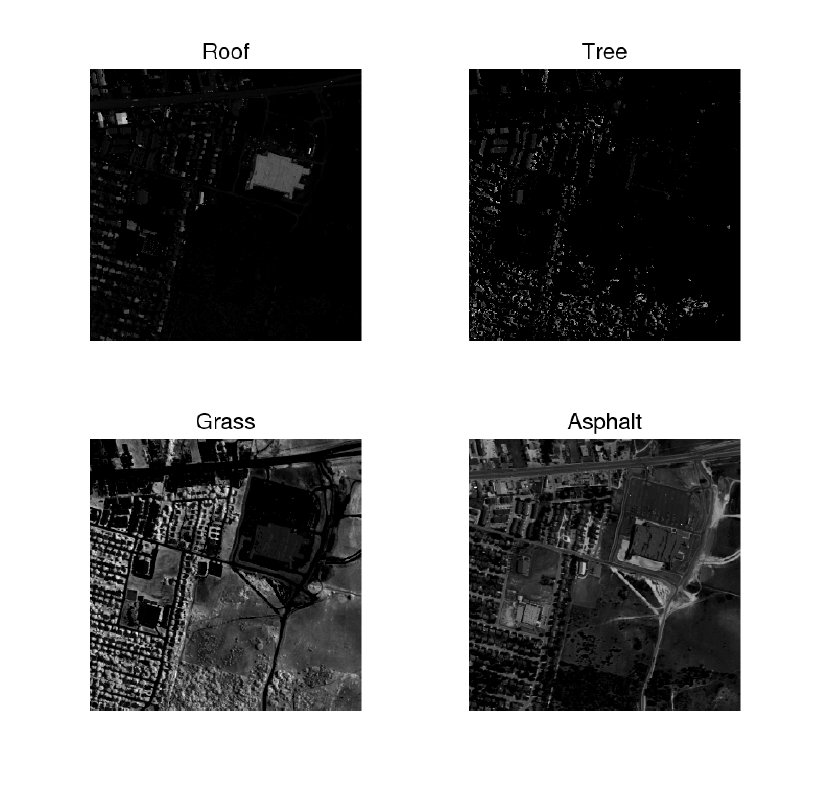}

	\caption{The abundance fraction maps of the proposed algorithm applied on HYDICE Urban dataset.}
	\label{HydiceAbundance}
\end{figure}

In \figurename{~\ref{realClustering}}, the results of FCM clustering on two real datasets is illustrated. The simulation results of spectral signatures have been shown in \figurename{~\ref{real}} and {~\ref{realhy}}. Also, $SAD$ performance metric of VCA-FCLS, $L_{1/2}$-NMF, distributed unmixing, sparsity constrained distributed unmixing and proposed method on the real datasets have been compared in \tablename{~\ref{table1}} and {~\ref{tablehy}}, the results of proposed algorithm are available in the last column and has the best $rmsSAD$ value. In \figurename{~\ref{CupriteAbundance}} and {~\ref{HydiceAbundance}} abundance fraction maps of the proposed method has been illustrated.


\begin{table}[htp]
	\centering
	\caption{
		The SAD and RE performance metric and their variance (in percent) of seven algorithms on AVIRIS Cuprite dataset, using VCA-FCLS initialization.
	}
	\begin{tabular}{c|c|c|c|c|c|c|c}
		\hline
		materials & VCA-FCLS & $L_{1/2}$-NMF & GLNMF & TV-RSNMF & Dist. & S. Dist. & Proposed Al.\\
		\hline \hline
		Sphene &0.3091 &0.2143 & 0.1913 &0.1583 & \textbf{0.1561} & 0.1673 & 0.1574$\pm$2.34\\
		\hline
		Nontronite & 0.2622 & 0.2518 &0.1842 &0.1803 & 0.1944 & 0.1743 &\textbf{0.1711}$\pm$1.89\\
		\hline
		KaolinSmect \#1 &0.2498 & {0.1653} &\textbf{0.1638}&0.1731& 0.2370 & 0.1741 & 0.1702$\pm$2.04\\
		\hline
		Montmorillonite & 0.2609 & 0.2318 &0.2184 &0.2159 &0.3571 & \textbf{0.2103} & 0.2248$\pm$2.54\\
		\hline
		Chalcedony & 0.1934& 0.1995 &0.1649 &0.1588 & 0.1603 & 0.1653 & \textbf{0.1437}$\pm$2.61\\
		\hline
		KaolinSmect \#2 & 0.3258 &\textbf{0.2542} &0.2594 &0.2576 &0.2873 & 0.2608 & 0.2596$\pm$1.87\\
		\hline
		Alunite &0.3601 &0.3458 &0.2841 &0.2551 &0.3813 & \textbf{0.2369} & 0.2417$\pm$2.09\\
		\hline
		Buddingtonite &0.2402 & 0.1693 &0.2068 &0.2034 &0.2514 & 0.1953 & \textbf{0.1643}$\pm$1.91\\
		\hline
		Muscovite & 0.3917 & 0.1584 &\textbf{0.1471} &0.1563 &0.4682 & {0.1537} & 0.1575$\pm$1.54\\
		\hline
		Andradite \#1 &0.2851 &0.3361 &0.3148 &0.2392 &\textbf{0.2132} & 0.2425 & 0.2337$\pm$2.67\\
		\hline
		Dumortierite & \textbf{0.2311}&0.2453 &0.2632 &0.2686 & 0.3381 & 0.2639 & 0.2519$\pm$2.98\\
		\hline
		Andradite \#2 &0.4492 & 0.3829 &0.3021 &0.3136 &0.3711 & 0.2854 &\textbf{0.2472}$\pm$3.87\\
		\hline\hline
		rmsSAD &0.3049 & 0.2562 &0.2317 &0.2207 &0.2998 & 0.2153 & \textbf{0.2064}$\pm$2.79\\
		\hline\hline
		RE & 0.0055 & 0.0050 & 0.0047 & 0.0038 & 0.0052 & 0.0035
		& \textbf{0.0029}$\pm$0.07\\
		\hline
	\end{tabular}
	\label{table1}
\end{table}
\begin{table}[H]
	\centering
	\caption{
		The SAD and RE performance metrics and their variance (in percent) of five algorithms on HYDICE Urban dataset, using VCA-FCLS initialization.
	}
	\begin{tabular}{c|c|c|c|c|c|c|c}
		\hline
		materials & VCA-FCLS&$L_{1/2}$-NMF & GLNMF & TV-RSNMF & Dist. & S. Dist. & Proposed\\
		\hline \hline
		Roof &0.4671 & 0.3461 &0.3486 &0.3327 &0.3831 & 0.3294 & \textbf{0.3289}$\pm$2.56\\
		\hline
		Tree & 0.2711 & \textbf{0.1492} &0.1673 &0.1572 &0.2052 & 0.1521 & 0.1496$\pm$3.34\\
		\hline
		Asphalt & 0.3077 & 0.2984 &0.2096 &\textbf{0.2054} &{0.2469} & {0.2118} & 0.2122$\pm$1.96\\
		\hline
		Grass & 0.2089 & 0.1461 &0.1283 &0.1249 &0.1344 & 0.1019 & \textbf{0.1014}$\pm$2.23\\
		\hline\hline
		rmsSAD &0.3279 & 0.2512 &0.2291 &0.2198 &0.2588 & 0.2161 &\textbf{0.2155}$\pm$2.47\\
		\hline\hline
		RE & 0.0134 & 0.0120 & 0.0112 & 0.0108 & 0.0131 & 0.0105
		& \textbf{0.0096}$\pm$0.11\\
		\hline
	\end{tabular}
	\label{tablehy}
\end{table}

\section{Conclusion}
This paper proposed the clustered multitask network scheme to solve SU problem. This new algorithm considered sparsity, clustering and neighborhood information. Simulation results on synthetic and real datasets illustrated preference of the proposed approach in comparison against previously published unmixing methods in terms of SAD, AAD and RE measures. Quantitatively, the proposed method improves RE metric for about 15 percent in AVIRIS Cuprite data experiment. Despite this performance improvement, the algorithm needs FCM clustering as a preprocessing stage that will increase computational complexity and run time of the algorithm. For future works, in this paper the FCM clustering method has been used, however using more efficient clustering methods to improve the clustered network can enhance the results of the proposed method.


\bibliography{report}   
\bibliographystyle{spiejour}   


\vspace{2ex}\noindent\textbf{Sara Khoshsokhan} received the B.Sc. degree in electrical engineering from Qom University of Technology in 2015, and the M.Sc. degree in communications system engineering from the Qom University of Technology in 2017. Her research interests include digital signal processing, pattern recognition and hyperspectral data analysis.

\vspace{2ex}\noindent\textbf{Roozbeh Rajabi} is an assistant professor at the Qom University of Technology. He received his BS and MS degrees in electrical engineering from the Iran University of Science and Technology and Tarbiat Modares University in 2007 and 2009, respectively, and his PhD degree in communications system engineering from the Tarbiat Modares University in 2014. His current research interests include hyperspectral data analysis, pattern recognition, and biomedical signal and image processing.

\vspace{2ex}\noindent\textbf{Hadi Zayyani} was born in Shiraz, Iran in 1978. He received the B.Sc., M.Sc. and Ph.D. all in Communications Engineering and all from Sharif University of Technology, Tehran, Iran. He is currently an assistant professor at the Faculty of Electrical and Computer Engineering, Qom University of Technology, Qom, Iran and also with School of Electrical and Computer Engineering, Shiraz University, Shiraz, Iran. His research interest include statistical signal processing, sparse signal processing, compressed sensing, adaptive filters and their applications.


\end{spacing}
\end{document}